\documentclass[sigplan,nonacm]{acmart}
\usepackage{url}
\usepackage{setspace}

\usepackage{enumitem}
\newlist{steps}{enumerate}{1}
\setlist[steps, 1]{label = Step \arabic*:}

\usepackage[T1]{fontenc}
\usepackage{lmodern}

\newcommand{\utterance}[1]{{\itshape\sffamily "#1"}}
\newcommand{\facet}[1]{{\scshape\sffamily #1}}
\newcommand{\tags}[1]{{\scshape\sffamily\lowercase{#1}}} 
\newcommand{\productcategory}[1]{{\scshape\sffamily\lowercase{#1}}}
\newcommand{\operator}[1]{{\scshape\sffamily\lowercase{#1}}}
\newcommand{\intentop}[1]{{\scshape\sffamily{#1}}}
\newcommand{\intentarg}[1]{{\scshape\sffamily\lowercase{#1}}}

\begin{document}
\title{ShopTalk: A System for Conversational Faceted Search}

\author{Gurmeet Manku, James Lee-Thorp, Bhargav Kanagal, Joshua Ainslie, Jingchen Feng,
Zach Pearson, Ebenezer Anjorin, Sudeep Gandhe, Ilya Eckstein
Jim Rosswog, Sumit Sanghai, Michael Pohl, Larry Adams, D. Sivakumar}
\affiliation{%
  \institution{Google}
  \country{United States}
}
\email{{manku, jamesleethorp, bhargav, jainslie, jingchenfeng}@google.com}
\email{{zpearson, eanjorin, srgandhe, ilyaeck, jrosswog, sumitsanghai, mpohl, leadams, siva}@google.com}

\renewcommand{\shortauthors}{Manku, et al.}

\begin{abstract}
We present ShopTalk, a multi-turn conversational faceted search system for Shopping that is designed to handle large and complex schemas that are beyond the scope of state of the art slot-filling systems. ShopTalk decouples dialog management from fulfillment, thereby allowing the dialog understanding system to be domain-agnostic and not tied to the particular Shopping application. The dialog understanding system consists of a deep-learned Contextual Language Understanding module, which interprets user utterances, and a primarily rules-based Dialog-State Tracker (DST), which updates the dialog state and formulates search requests intended for the fulfillment engine. The interface between the two modules consists of a minimal set of domain-agnostic ``intent operators,'' which instruct the DST on how to update the dialog state. ShopTalk was deployed in 2020 on the Google Assistant for Shopping searches.
\end{abstract}




\maketitle

\section{Introduction}
\label{section:introduction}

Faceted Search is a classic paradigm for object search in a collection of objects. For example, Amazon, eBay and Yelp use Faceted Search to help us find the right product or the right business location. Such an experience offers two knobs for users to specify what they are searching for: a text query and facet selections.
\begin{itemize}
    \item 
        Text queries are keyword based, for example, \utterance{red Nike shoes.} In recent years, especially with the rise of mobile devices and breakthroughs in ASR (Automatic Speech Recognition) and NLU (Natural Language Understanding), longer and more natural language queries have risen in prominence, for example, \utterance{show me some red Nike shoes please.}
    \item
        Facets are object attributes such as \facet{Brand} and \facet{Price}. The values that can be selected within a facet are known as tags. For example, \tags{Nike} and \tags{\$50} are tags for \facet{Brand} and \facet{Price} facets, respectively. Facets are usually presented to a user using dropdowns, checkboxes, radio buttons and text fields for range specification. In Shopping schema, a special facet called \facet{ProductCategory} (with tags like \tags{Shoes} and \tags{Televisions}) helps us distinguish between disparate object collections.
\end{itemize}

A Faceted Search experience is particularly helpful when the user knows what they are looking for and they are able to express their needs as a combination of text query and facet selections.

\subsection{Problem Definition}

The research problem we tackle in this paper is: 
\begin{quote}
  {\em How may we enable classic Faceted Search in a chatbot-style conversational system where a user drives the conversation by expressing a series of preferences?}
\end{quote}
We assume the existence of (a) a Shopping schema with facets and tags, (b) a search backend to fulfill user requests from query and facet constraints, and (c) query logs for the existing, non-conversational Faceted Search experience. Figure~\ref{fig:utterance_sequence} shows an example utterance sequence that ShopTalk is able to support.

\begin{figure}
    \begin{enumerate}
        \item
            \utterance{Show me some \underline{Nike} \underline{women’s} \fbox{shoes} please.}
        \item
            \utterance{Do you have anything in \underline{red}?}
        \item
            \utterance{How about \underline{pink}?}
        \item
            \utterance{Actually, almost \underline{any color} will do; just make sure it’s \underline{not white}.} \\
            <scrolls through products on display>
        \item
            \utterance{Hmmm… let’s \underline{also} see \underline{Adidas}.}
        \item
            \utterance{Okay, it \underline{doesn’t have to be Adidas} but I want ones that are \underline{good for running}.}
        \item
            \utterance{Something that \underline{protects my feet in heavy rain}.}
        \item
            \utterance{Do you have anything \underline{less than a hundred bucks}?}
        \item
            \utterance{Anything even \underline{cheaper}?} \\
            <clicks on a product>
        \item
            \utterance{\underline{Size 9}.} \\
            <selects product >
        \item
            \utterance{I want to buy some \underline{red} \fbox{socks} too.}
    \end{enumerate}
    \caption{A complex utterance sequence parsed by ShopTalk. Utterances (1) and (11) trigger product category switches. Utterances (1), (4) and (6) express multiple preferences in the same utterance. Utterance (4) expresses a negative preference. Utterances (8) and (9) showcase range-oriented preferences and nudges over a numeric attribute. Almost all utterances refer to tags but utterances (4) and (10) refer to facets \facet{Color} and \facet{Size} by their name. Utterance (7) highlights a long text span that maps to tag \tags{waterproof}.}
    \label{fig:utterance_sequence}
\end{figure}

\subsection{Challenges}
\label{subsection:challenges}

Understanding natural language utterances accurately is a challenging research problem in its own right. We argue that parsing Shopping utterances is especially so, owing to the {\em large}, {\em complex} and {\em incomplete} nature of the shopping schema.

\paragraph{\bf Shopping schema is large:} A web-scale shopping schema like Google product taxonomy has over 6,000 product categories~\cite{GoogleProductTaxonomy}. AliExpress taxonomy contains thousands of product categories, and a single category, Sports and Entertainment, has over 8,900 unique attributes~\cite{xu-etal-2019-scaling}. Such schemas are much larger than schemas tackled by state-of-the-art slot filling dialog systems~\cite{Gao2019,Goel2019,Chao2019}. For comparison, the MultiWOZ~\cite{Budzianowski2018,Ramadan2018} and DSTC8~\cite{Rastogi2019} task-oriented challenges consist of only thousands of attributes across the entire schema.

\paragraph{\bf Shopping schema is complex:}
\begin{enumerate}
    \item 
        Facets come in different types. Specifically, facets may be boolean-valued (e.g., \tags{waterproof}), numeric (e.g., \facet{Price}: \tags{\$20} to \tags{\$200}), discrete ordered (e.g., \facet{Size}: \tags{XS}, \tags{S}, \tags{M}, \tags{L}, \tags{XL}, \tags{XXL}) or unordered (e.g., \facet{TypeOfCamera}: \tags{Disposable}, \tags{Point \& Shoot}, \tags{SLR}, \tags{Medium Format}). Users specify preferences over these types in a variety of ways like \utterance{it must be \underline{waterproof}} or \utterance{i’m not looking for a \underline{throwaway camera}}. Types that may be ordered allow users to specify ranges and sort ordering, leading to utterances like \utterance{something \underline{less than \$200}} $\rightarrow$ \utterance{show me something \underline{cheaper}} $\rightarrow$ \utterance{show me the \underline{cheaper ones first}}.
    \item
        Within a product category, most tags belong to a unique facet. However, it's possible that the same tag belongs to multiple facets. For example, a numeric tag like \tags{5} may represent the \facet{size} or \facet{volume} or \facet{length} of that product category. A parser is faced with the challenge of resolving such ambiguities.
    \item 
        Some facets can have multiple types. For example, the \facet{Size} facet for product category \productcategory{Shoes} may be both numeric (\tags{6}, \tags{7}, \tags{8}, …, \tags{14}) and non-numeric (\tags{S}, \tags{M}, \tags{L}, \tags{XL}) at the same time.
\end{enumerate}

\paragraph{\bf Shopping schema is incomplete:} Web-scale Shopping schemas are inherently dynamic. They are constantly updated as new products and their associated tags \& facets are introduced every day. Also, Shopping schemas are rich with respect to tags and facets for popular product categories (the “head” of the schema) but impoverished for less popular categories (the “tail” of the schema).

\paragraph{\bf The ungrounded span problem:} An ungrounded span in a user utterance is a token sequence referring to a tag that is missing in the schema. Examples:
\begin{enumerate}
    \item
        \utterance{show me women's shoes without any \underline{ankle straps}}
    \item  
        \utterance{i want to buy \underline{lemon-scented} hand soap} $\rightarrow$ \utterance{no, i actually prefer \underline{lavender}}
\end{enumerate}
where \utterance{ankle strap}, \utterance{lemon-scented} and \utterance{lavender} may refer to tags missing in Shopping schema. In example (2), a smart dialog understanding system would replace the preference for \utterance{lemon-scented} by \utterance{lavender} even though both are missing in the schema.

\paragraph{\bf Synonymization problem:} Users often do not conceptualize facets and tags in terms of names chosen by Shopping schema builders. In fact, the schema terminology may not be intuitive to a consumer because it is chosen by product manufacturers or identified through a combination of manual curation and machine learning. Moreover, not all consumers think of a facet or tag using the same words to describe product attributes. For example, both \utterance{does not get wet} and \utterance{protects my feet in heavy rain} are synonyms for tag \tags{Waterproof}.

\paragraph{\bf Utterances are contextual:} To parse the latest utterance, we may need a context established by one or more previous utterances. Consider \utterance{i want to buy a tv} $\rightarrow$ \utterance{something with 5 ports} $\rightarrow$ \utterance{can we increase that?} In the second and third utterances, the user did not (and need not) specify the product category (\productcategory{Televisions}). We need to detect that neither of these utterances is a product category switch. The third utterance has no tag or facet; it's a request to nudge some numeric facet. In order to parse this successfully, we need context established by the previous utterance, namely the facet \facet{NumberOfPorts}.

\paragraph{\bf Negative preferences:} Users may wish to rule out portions of the product space with utterances like \utterance{I don’t want Nike}, or \utterance{I hate red}, or \utterance{don’t show me front-loading dishwashers.} Such negative preferences are not supported by classic faceted search interfaces, but they are natural and common in spoken language interfaces.

\paragraph{\bf Multiple preferences in an utterance:} For example, \utterance{show me red Nike shoes} specifies three preferences: (\facet{ProductCategory} = \productcategory{Shoes}) \operator{AND} (\facet{Brand} = \tags{Nike}) \operator{AND} (\facet{Color} = \tags{Red}). A follow-on utterance like \utterance{don’t show me blue; I like red} specifies two preferences: \operator{NOT} (\facet{Color} = \tags{Blue}) \operator{AND} (\facet{Color} = \tags{Red}).

\subsection{Research Contributions}
\label{subsection:research_contributions}

\paragraph{\bf Real world deployment:} Our system was integrated with Google Assistant and deployed in March 2020. Integration was challenging due to the inherent nature of Google Assistant as a large-scale conversation system that multiplexes multiple systems like Google Search, Google Maps, Hotel Search and Shopping.

\paragraph{\bf System design:} We implemented a 3-stage pipeline:
\begin{itemize}
    \item[\bf\sc parse:] A CLU (Contextual Language Understanding) module with SOTA seq2seq models for parsing the latest user utterance into a structured, human-understandable interpretation. See Section~\ref{section:clu} for details.
    \item[\bf\sc merge:] A DST (Dialog State Tracking) module for merging the interpretation of the latest utterance with the interpretations of previous utterances to derive cumulative dialog state. See Section~\ref{section:dst} for details.
    \item[\bf\sc fulfill:] A Fulfillment module to respond to user utterances by displaying products that match a text query and cumulative facet preferences, both of which are derived from dialog state. See Section~\ref{subsection:fulfillment} for details.
\end{itemize}

\paragraph{\bf Domain agnostic design:} The CLU-DST system is designed to work for conversational faceted search in multiple domains / verticals with a typical faceted search schema. A unique feature of our design is that the CLU sends domain-agnostic {\em intent operators} to the DST to update dialog state for complex schemas (see Section~\ref{section:system_overview}).

\paragraph{\bf Training data collection:} In the absence of an existing conversational faceted search system, we could not leverage logs of existing dialog sessions or dialog-oriented training data for developing CLU \& DST modules. So we developed two approaches for training data collection:
\begin{itemize}
    \item 
        Innovative strategies for collecting a minimal set of utterances from human raters. In fact, with only a few thousand such utterances, we were able to build a CLU-DST system for a web-scale Shopping schema. See Sections~\ref{section:clu} and~\ref{section:dst}.
    \item 
        Synthetic dialog generation by leveraging faceted search journeys in existing GUI-based Shopping systems. We explain this process in Section~\ref{section:clu}.
\end{itemize}

\paragraph{\bf Support for complex utterances \& preferences:} Our CLU \& DST modules support all types of facets. Negative preferences and multiple preferences specified in the same utterance are also supported. We also address the synonymization problem by successfully mapping spans like "water resistant" to \tags{waterproof} tag.

\paragraph{\bf The ungrounded span problem:} Our CLU \& DST modules tackle two thorny problems arising from Shopping schema incompleteness:
\begin{itemize}
    \item 
        {\em The Missing Tags Problem}: Our CLU module is capable of discovering tags from user utterances on-the-fly even if that tag was not part of Shopping schema. We call such tags “ungrounded spans” because these are text spans that the CLU was not able to map to a tag known to the schema. We also discover negative preferences for ungrounded spans.
    \item 
        {\em The Missing Tag Aggregations Problem}: Our DST module is capable of merging user preferences associated with ungrounded spans with user preferences already enunciated by the user so far (even though tags corresponding to ungrounded spans are not part of the schema). See Section~\ref{subsection:classifier_model} for an exposition of the learned model used by the DST for handling ungrounded spans.
\end{itemize}


\section{Related Work}
\label{section:related_work}

In recent years, multiple approaches have been proposed for task-oriented dialog state tracking. An incomplete, short list of such approaches includes pure rule-based approaches~\citep{Wang2013,yan2017}, Bayesian networks~\citep{Thomson2010}, Conditional Random Fields (CRF)~\citep{Lee2013}, recurrent neural networks~\citep{Henderson2014,DBLP:conf/aaai/GongLZOLZZDC19}, end-to-end memory networks~\citep{Perez2016}, pointer networks~\citep{Xu2018}, embedding-based approaches~\citep{Mrksic2016, Ren2018}, hybrid approaches~\citep{Goel2019} and BERT-based approaches~\citep{Chao2019}. Please see Weld et al.~\cite{weld2021survey} for a comprehensive survey.
We draw attention to two exemplary works to contrast with our approach.

Related to e-commerce search, our work is most similar to Yang et al~\citep{Yang2018}, who pose the problem of dialog state tracking as a query tracking problem --- after each user utterance, their system updates a single query that is sent to the fulfillment backend. There are two major contrasts with our work:
\begin{itemize}
    \item 
        {\bf Query state vs structured dialog state}: The query-based approach by Yang et al~\cite{Yang2018} relies on the assumption that the word order in queries is unimportant. However, in deploying our system, we found that the ability to selectively fulfill user preferences, based on a recency bias, was required in order to serve a non-zero number of results in multi-turn conversations\footnote{We also found that representing the user’s preferences as a query limits a user’s visibility into the system’s conversational understanding.}.
    \item 
        {\bf End-to-end vs modular systems:} Our design separates intent parsing from dialog state updates. In contrast, Yang et al~\citep{Yang2018} use an end-to-end text based attention model. Such models require large amounts of training data to handle complex transitions (e.g., involving synonyms); they can be difficult to understand and debug in production services.
\end{itemize}

Most existing task-oriented dialog state tracking systems assume a fixed vocabulary of tags and facets. Gao et al~\cite{Gao2019} attempt to overcome the limitations of fixed-vocabularies by adopting a neural reading comprehension approach. They tackle the dialog state tracking problem by making three sequential decisions based on the existing dialog state and the incoming utterance: (i) Should this slot from the existing dialog state be carried over the new dialog state? (ii) What slot type should the model update? (iii) What slot value should be copied to the selected slot type? This sequential or gated decision making setup is a relatively common and successful approach; see also~\cite{Wu2019} for example. Gao et al~\cite{Gao2019} base their model on BERT and use the open benchmark MultiWOZ dataset~\cite{Budzianowski2018,Ramadan2018}. However, their “zero-shot” attribute results are necessarily limited by the relative size of the MultiWOZ multi-turn dialog, which has only 24 facets and 4,510 tags. 

Many more multi-turn dialog datasets have emerged in recent years, generally relying on a Wizard of Oz collection setup~\cite{Wen2016,Asri2017}. These datasets typically span multiple domains and are geared towards general virtual assistant settings. Nevertheless, even the largest of these datasets (at the time of writing), namely the Schema-Guided Dialogue dataset~\cite{Rastogi2019}, contains significantly fewer facets (214) and tags (14,139) than ours. In contrast, a web-scale Shopping schema like Google product taxonomy has over 6,000 product categories~\cite{GoogleProductTaxonomy}. AliExpress taxonomy has thousands of product categories; a single category -- Sports \& Entertainment -- has over 8,900 unique attributes~\cite{xu-etal-2019-scaling}. Moreover, such schemas are incomplete and dynamic; new tags and facets are introduced almost daily.


\section{System Overview}
\label{section:system_overview}

\begin{figure}
    \centering
    \includegraphics[width=\columnwidth]{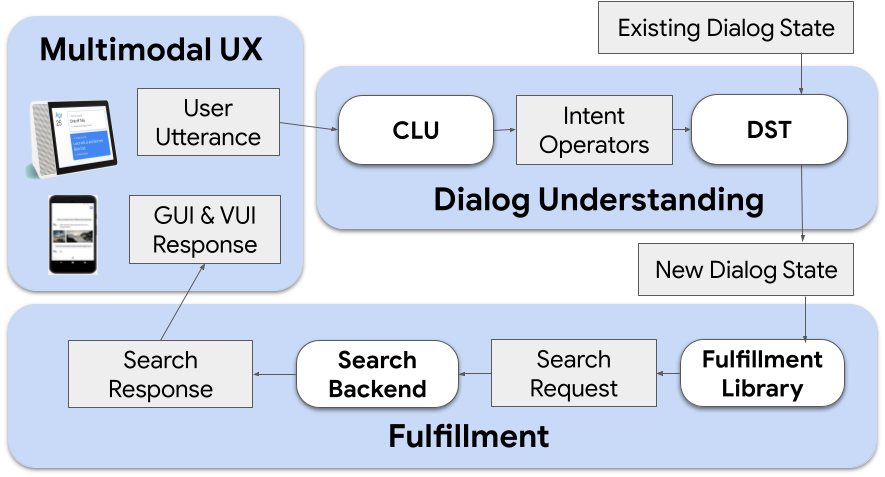}
    \caption{ShopTalk system architecture.}
    \label{figure:system_diagram}
\end{figure}

Figure~\ref{figure:system_diagram} is a block diagram for ShopTalk. The CLU module parses a user utterance into intent operators (defined in Section~\ref{subsection:intent_operators}). The DST module uses these intent operators to update an existing dialog state to produce a new dialog state (defined in Section~\ref{subsection:dialog_state_representation}). The Fulfillment module computes a query and facet constraints from the new dialog state. Products retrieved by the Fulfillment module are shown to the user.

\subsection{Dialog State Representation}
\label{subsection:dialog_state_representation}
Dialog state representation is a key design decision in dialog systems. We chose a structured format: an unordered set of predicates over facets ("slots”) and tags ("values”) to represent cumulative search preferences specified by a user so far. Some of the tags in dialog state may be ungrounded spans. An example:
\begin{quote}
  (\facet{ProductCategory} = \productcategory{Shoes})  \\
  \operator{AND} ((\facet{Size} = \tags{10}) \operator{OR} (\facet{Size} = \tags{11}))  \\
  \operator{AND} (\facet{Color} $\neq$ \tags{Red}) \operator{AND} (\facet{Color} $\neq$ \tags{Blue})  \\
  \operator{AND} \utterance{square heels} \\
  \operator{AND} \facet{SortOrder} = (\facet{Price}, \operator{ASCENDING})
\end{quote}


Maintaining dialog state in a structured format, as opposed to just a query summary of search preferences, has two primary benefits:
\begin{enumerate}
    \item 
        Dialog state may be echoed back to the users in a human friendly format for {\em grounding}, which helps the user establish trust that their utterance was understood by the system.
    \item
        Dialog state is portable and domain agnostic. Any specific application only needs an adapter module to convert dialog state into their fulfillment language. For Shopping, we convert dialog state to a combination of search query and facet restricts.
\end{enumerate}

\subsection{Intent Operators}
\label{subsection:intent_operators}

The CLU parses user utterances into intents. An intent is defined by an intent operator and its associated arguments. See Figure~\ref{figure:intent_operators} for examples. Intent operators may be thought of as DST operators -- they instruct the DST on how to update the existing dialog state. We designed intent operators carefully as a finite, minimal and complete basis for all possible DST $\rightarrow$ DST transitions that we identified. We distinguish intent operators by whether they act on facets or tags in dialog state.



{\bf Tag-level intent operators} are used to select or deselect tags:
\begin{itemize}
    \item
        \intentop{SetValueOp}($\langle$tag$\rangle$, $\langle$predicate-type$\rangle$, $\langle$inclusivity$\rangle$) is the richest (and most overloaded) of intent operators. It is also the most common intent operator in real world dialog. \intentop{SetValueOp} represents utterances such as \utterance{show me size 8}, \utterance{show me size 8 also}, \utterance{show me size 8 only} and \utterance{size 8 or more}. Argument $\langle$inclusivity$\rangle$ has 3 possible values: \operator{INCLUSIVE} (\utterance{also}), \operator{EXCLUSIVE} (\utterance{only}) and \operator{UNDEFINED} (unspecified). Argument $\langle$predicate-type$\rangle$ has 6 possible values: \operator{EQUALS}, \operator{NOT\_EQUALS}, \operator{LESS\_THAN}, \operator{LESS\_EQ}, \operator{GREATER\_THAN} and \operator{GREATER\_EQ} to represent $=$, $\neq$, $<$, $\leq$, $>$ and $\geq$, respectively. Predicate types for $<$, $\leq$, $>$ and $\geq$ express ranges which are valid only for facets of type numeric and categorical ordered. If $\langle$predicate-type$\rangle$ expresses ranges, the dialog state update requires merging of tag ranges. Otherwise, dialog state update amounts to simply appending ($\langle$tag$\rangle$, $\langle$predicate-type$\rangle$) to dialog state. Additionally, if $\langle$inclusivity$\rangle$ equals \operator{EXCLUSIVE}, then all other tags in the same facet as $\langle$tag$\rangle$ are cleared.
    \item
        \intentop{ClearValueOp}($\langle$tag$\rangle$) represents utterances such as \utterance{i don’t care if it’s red or not}; the DST forgets $\langle$tag$\rangle$ by clearing it out. Note that this is different from \utterance{i don't want red}, which would be a \intentop{SetValueOp}(\tags{Red}, \operator{NOT\_EQUALS}).
\end{itemize}

{\bf Facet-level intent operators} trigger facet-level adjustments:
\begin{itemize}
    \item 
        \intentop{ClearFacetOp}($\langle$facet$\rangle$) clears all tags and predicates for $\langle$facet$\rangle$, e.g., \utterance{any color will do} clears out any predicates associated with facet \facet{Color}.
    \item 
        \intentop{ClearAllFacetsOp}() clears dialog state completely by removing all tags and predicates for all facets, e.g., \utterance{let’s start over}.
    \item 
        \intentop{NudgeFacetOp}($\langle$facet$\rangle$, $\langle$nudge-direction$\rangle$) where $\langle$nudge-direction$\rangle$ may be \operator{POSITIVE} or \operator{NEGATIVE} adjusts the tag for an ordered facet up or down, e.g., \utterance{i want something larger}.
    \item 
        \intentop{OrderByOp}($\langle$facet$\rangle$, $\langle$sort-direction$\rangle$) specifies the sort order for fulfillment, e.g., \utterance{show me the cheapest}.
\end{itemize}

Note that a single user utterance may give rise to multiple intent operators. For example, \utterance{i don’t care about the color but i want size 10} parses into two intent operators. We also emphasize that the intent operators defined above do not incorporate any Shopping specific terminology and are therefore domain agnostic.

\begin{figure}[th]
    \small
    \setstretch{1.2} 
    \begin{tabular}{p{\dimexpr 0.5\linewidth - 2\tabcolsep}p{\dimexpr 0.5\linewidth - 2\tabcolsep}}
        $\qquad$ {\bf User utterance}
        & $\quad$ {\bf Intents (Output of CLU)} \\
        (1) \utterance{Show me some Nike shoes}
        & \intentop{SetValueOp}(\operator{EQUALS}, \tags{Nike})\\
        (2) \utterance{Something for running}
        & \intentop{SetValueOp}(\operator{EQUALS}, \tags{running})\\
        (3) \utterance{Adidas ones too please}
        & \intentop{SetValueOp}(\operator{EQUALS}, \intentarg{INCLUSIVE}, \tags{Adidas})\\
        (4) \utterance{Orange is okay but I don’t want pink}
        & \{ \intentop{SetValueOp}(\operator{EQUALS}, \tags{orange}); \intentop{SetValueOp}(\operator{NOT\_EQUALS}, \tags{pink}) \}\\
        (5) \utterance{Do you have anything in razmatazz?}
        & \intentop{SetValueOp}(\operator{EQUALS}, \utterance{razmatazz})\\
        (6) \utterance{Actually any color is OK}
        & \intentop{ClearFacetOp}(\facet{Color})\\
        (7) \utterance{Size 9}
        & \intentop{SetValueOp}(\operator{EQUALS}, \tags{Size9})\\
        (8) \utterance{Show me something bigger}
        & \intentop{NudgeFacetOp}(\intentarg{INCREASE}, \facet{Size})\\
        (9) \utterance{It doesn't have to be black}
        & \intentop{ClearValueOp}(\tags{black})\\
        (10) \utterance{Do you have anything less than fifty bucks?}
        & \intentop{SetValueOp}(\operator{LESS\_THAN}, \tags{Price$50$})\\
        (11) \utterance{start over}
        & \intentop{ClearAllFacetsOp}()\\
    \end{tabular}
    \caption{Examples of user utterances mapped to intents. In utterance (5), \utterance{razmatazz} is an ungrounded span. Utterance (8) needs the context of utterance (7) for interpretation.}
    \label{figure:intent_operators}
\end{figure}

\subsection{Separation of Intent Parsing and Dialog State Tracking}

Decomposition of the end-to-end dialog understanding problem into intent parsing (CLU) and dialog state tracking (DST) is motivated by the scale and complexity of the Shopping schema. The number of possible transitions from old dialog state to new dialog state is potentially of quadratic complexity in the number of unique tags in the schema. Factoring dialog state transitions into a small-sized ($O(1)$) set of intent operators simplifies training data collection dramatically. Instead of collecting training examples for all possible dialog state transitions, we just need a representative sample for each intent operator. For example, rather than collect training data examples of the form \utterance{i want red only}, \utterance{i want blue only}, \utterance{i want Nike only}, \ldots, it suffices to identify an intent operator for the canonical utterance \utterance{i want $\langle$tag$\rangle$ only} and then collect training examples for a sample of $\langle$tag$\rangle$ values. Note that canonical utterances are product category and domain agnostic.

\subsection{Annotated Spans Vs Ungrounded Spans}

A span is a token sequence in an utterance. A span can be annotated if it is recognized by the schema as a tag or a facet. Such annotations are helpful to both the CLU and DST. They constitute a strong signal that the span is relevant, and that the CLU should copy the span to construct a facet or a tag for a relevant intent operator; see Section~\ref{section:clu}. For the DST, updating the dialog state for an intent operator with an annotated span reduces to a facet (slot) filling problem; see Section~\ref{section:dst}. For example, if we know from our schema that \tags{red} is a \facet{Color}, then the DST knows to fill in the \facet{Color} slot with \tags{red}.

A web-scale Shopping schema is inherently incomplete. What happens if the spans cannot be mapped to tags or facets in the schema? So-called {\em ungrounded spans} are more difficult for the CLU to detect, and also cannot be mapped directly to facets by the DST. The ungrounded span problem requires special care in both the CLU and DST; see Sections~\ref{section:clu} and~\ref{section:dst}, respectively.

\subsection{Fulfillment}
\label{subsection:fulfillment}
The Fulfillment module converts dialog state into a combination of search query and facet restricts for a state of the art Faceted Search engine like Apache Solr on Lucene~\cite{ApacheSolr, ApacheLucene}. Facet restricts are easy to compute from facet predicates. We construct the search query by combining all ungrounded spans and their associated predicates with a canonical phrase for the product category in dialog state; such canonical phrases are part of Shopping schema. The search query and facet restricts are sent together to a Shopping search engine to retrieve matching products.



\section{Contextual Language Understanding}
\label{section:clu}

In this section, we present the design of our CLU module which parses a user utterance into a set of intents, based on the context of the dialog. See Section~\ref{subsection:intent_operators} for a list of intent operators.

The task of parsing utterances into intents is commonly known as the semantic parsing problem~\cite{jia-liang-2016-data} which is well studied in literature. In recent years, several modeling approaches have been proposed~\cite{jia-liang-2016-data,shaw-etal-2019-generating,shaw-etal-2018-self,dong-lapata-2018-coarse,dong-lapata-2016-language,iyer-etal-2017-learning}. Inspired by the success of sequence-to-sequence models~\cite{dong-lapata-2018-coarse,dong-lapata-2016-language,zhang-etal-2019-editing,lin2019grammarbased}, we chose to formulate our parsing problem as a sequence-to-sequence task.

Examples of input-output pairs for the CLU are listed in Figure~\ref{figure:intent_operators}. Note that the CLU output is not a single intent but a {\em sequence of intents} to accommodate multiplicity of user preferences --- see utterance (4) in Figure~\ref{figure:intent_operators}, for example. Also, context is important. In utterance (2) in Figure~\ref{figure:intent_operators}, the user simply specifies \utterance{running} without specifying that \facet{ProductCategory} is \facet{Shoes}, which needs to be inferred from the first utterance.

\subsection{Challenges}

\subsubsection*{\bf Training data collection:}
For web-scale schemas~\cite{GoogleProductTaxonomy,xu-etal-2019-scaling} with millions of unique tags, the number of unique intents is at least O(100M) due to combinatorial explosion: Our intents are parameterized by $6$ intent operator types, $8$ predicate types, $3$ inclusivity types, and over a million unique tag values (see Section~\ref{section:introduction}). Thus collecting utterances for all possible intents is infeasible; utterances for multiple intents (e.g., utterance (4) in Figure~\ref{figure:intent_operators}) make it even more challenging.

\subsubsection*{\bf Handling ungrounded spans:}
Another challenge pertains to ungrounded spans defined in Section~\ref{section:introduction}. The CLU model needs to recognize such phrases and also predict the right intent operators (e.g., utterance (4) in Figure~\ref{figure:intent_operators}).
Note that for our target search application, fulfillment is feasible even for ungrounded spans, given access to a text search backend.
Dealing with ungrounded spans also leads to challenges with dialog state tracking which we will discuss in Section~\ref{section:dst}.

\subsection{\bf Training Data}
We develop three strategies to collect training data.

\subsubsection*{\bf Dataset D1 from raters:}
Collecting utterances for every intent is infeasible for O(100M) intents. However, observe that expressing preference for \tags{Nike} shoes is quite similar to expressing preferences for \tags{running} shoes --- see utterances (1) and (2) in Figure~\ref{figure:intent_operators}. This is because both utterances (1) and (2) map to the same {\em intent signature}, i.e., an intent that is not yet bound to a specific facet or tag (\intentop{SetValueOp} \operator{Equals} for this example). Based on this observation, we develop a strategy to collect utterances for every possible intent signature, for a small number of attributes. The total number of unique intent signatures is around $100$ and is much more manageable. We develop techniques to generalize to other attributes (Section~\ref{section:contextual_annotations}) and ungrounded spans (Section~\ref{section:clu_architecture}). We select $10$ attributes from each of $6$ categories (a mix of discrete, numeric and boolean attributes). For each resulting intent, we instruct the raters to specify at least 5 utterances (with an emphasis on diversity). For example, given the intent \intentop{SetValueOp}(\facet{Color}, \operator{NOT\_EQUALS}, \tags{Red}) for shoes, we obtain example utterances such as \utterance{I don't like red}, \utterance{don't show me red shoes}, \utterance{I hate red}, etc. To handle the complexity of the synonymization problem for popular facets like \facet{price}, we collect additional utterances. For example, note that prices can be expressed quantitatively (e.g., \utterance{under \$50}) or qualitatively (\utterance{something cheap}).

\subsubsection*{\bf Dataset D2 from query logs:}
As previously observed by Yang et al~\cite{Yang2018}, search engine logs are a large-scale source of user data for faceted search. However, two challenges emerge: First, queries are much less verbose than user utterances on the assistant; Second, we still need to annotate each query with the right intent. To leverage search engine logs for training, we first select frequently occurring queries, i.e., those that are queried at least $100$ times and then filter to long product-seeking queries. To automatically annotate the intent for each query, we further choose two subsets of queries:
\begin{itemize}
    \item 
        {\bf Fully parsed queries}: All terms in such queries belong to Shopping schema. We can easily label such queries with their intents. Fully parsed queries are a rich source of utterances with multiple intents.
    \item
        {\bf Well understood regular expressions}: these are queries that match a whitelist of carefully selected regular expressions. For example, using the regular expression \utterance{\$category that are good for \$span}, we can match queries such as \utterance{shoes that are good for back pain}. For such queries, we can construct the intent, namely \intentop{SetValue}(\operator{EQUALS}, \$span). The advantage with this approach is that we can construct training examples that contain unknown attributes captured by \$span. We can also obtain negations using templates like \utterance{\$category without \$span} (e.g., \utterance{shirts without stripes}).
\end{itemize}
Since search is a single-turn experience, all the utterances that we obtain correspond to a \intentop{SetValueOp}. Using query logs, we obtained over $100K$ training examples for the model.

\subsubsection*{\bf Dataset D3 from grammars:}
\noindent Context-free grammars can be used to generate user utterances as shown in Yang et al.~\cite{Yang2018}. To generate a realistic grammar, we leveraged the rater utterances from D1. For each intent signature, we collected applicable rater utterances, {\em delexicalized} the attribute name (\utterance{show me red ones} $\rightarrow$ \utterance{show me \$tag ones}) and heuristically induced a grammar that would generate much of the delexicalized utterances; see Figure~\ref{figure:grammar} for an example for \intentop{SetValueOp} \operator{NOT\_EQUALS}.

Next, we instantiated the grammar to generate all possible templates (without binding to specific tag values) yielding around $500K$. Then, we sample tag values from our schema to produce several million utterances for the CLU model. While rater utterances and user queries correspond to high training quality data, not all utterances obtained from grammars are fluent, therefore these were given a lower weight while training.

We observed the best model quality when we used all three datasets D1, D2 and D3 for training. D3, the dataset generated based on grammars gave a substantial boost to the performance of the model, despite it being lower quality. 

\begin{figure}[tb]
    \textbf{Grammar for} \_NegationUtterance \\
    \medskip
    \begin{tabular}{rl}
        \_NegationUtterance $\rightarrow$
        & (\_IDontWant) (\_Condition) | \\
        & (\_IWant) (\_NotCondition) \\
        \_IDontWant $\rightarrow$
        & "i (\_Dont) want" | "i (\_Dont) like" | \\
        & "i wouldn't like" | "i dislike" | \\
        & "i hate" | "i (\_Dont) want to see" \\
        \_IWant $\rightarrow$
        & "i want" | "i like" | "show me" \\
        \_Dont $\rightarrow$
        & "don't" | "do not" \\
        \_Condition $\rightarrow$
        & "$\langle$\tags{tag}$\rangle$" \\
        \_NotCondition $\rightarrow$
        & "no $\langle$\tags{tag}$\rangle$"
    \end{tabular}

    \medskip
    \par\noindent\rule{\columnwidth}{0.4pt}
    \textbf{Example utterances for $\langle$\tags{tag}$\rangle$ = \utterance{red}.} \\
    \medskip
    \utterance{i don't want red},
    \utterance{i do not want red},
    \utterance{i hate red},
    \utterance{i want no red},
    \utterance{i wouldn't like red},
    \utterance{i don't want to see red},
    \utterance{i dislike red}, \ldots
    \caption{Grammar-based training data generation.}
    \label{figure:grammar}
\end{figure}

\subsection{Contextual Annotations}
\label{section:contextual_annotations}

Our training data does not contain all attributes from the schema; in order to help the model generalize to all attributes, we provide additional features in the form of text spans in the utterance that correspond to known attributes. The presence of such an annotation on a text span is a signal to the model which indicates it is important to the utterance, even if it was not seen in the training data.

To compute annotations, we first need to predict the product category of interest to the user. We rely on an existing classifier for this task, applied to the utterance. However, users tend not to repeat the product category in follow-on utterances, in which case we reuse the category from a previous utterance in the dialog (unless we detect a category switch, see Section~\ref{section:deployment}).
Subsequently, we match all known tags (and their curated synonyms) in the product category to the query in order to determine the relevant annotations. Annotations are featurized using a graph encoding, as described below.

\subsection{Model Architecture:}
\label{section:clu_architecture}
We follow the model architecture outlined by Shaw et al.~\cite{shaw-etal-2019-generating, shaw-etal-2018-self}. It is based on the well known Transformer~\cite{NIPS2017_3f5ee243} encoder-decoder model, with support for handling {\em graph structured inputs} and uses a {\em copy mechanism} in the decoder. We briefly describe these below, and we also introduce a novel annotation dropout scheme to improve generalization for ungrounded spans.

The input to the transformer encoder is the user utterance along with its annotations, and the output from the decoder are the intents. The input text is tokenized into wordpieces~\cite{devlin-etal-2019-bert}. The annotations are treated as a special token and featurized purely by their type. The names of the intent operators and their parameters are added to the decoder vocab.

\begin{figure}[tb]
    \includegraphics[width=\columnwidth]{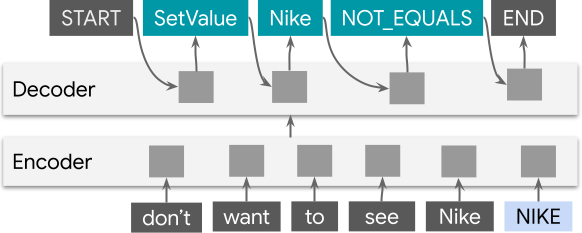}
    \caption{The CLU Transformer. Input: user utterance --- tokenized and annotated with known attributes (e.g., NIKE in the blue box). Output: intent shown on top. The text token `Nike' is copied from input to output.}
    \label{figure:seq2seq_model}
\end{figure}

\subsubsection*{\bf Graph inputs:}
Shaw et al.~\cite{shaw-etal-2019-generating, shaw-etal-2018-self} represent the input utterance and corresponding annotations using a graph. Nodes in the graph correspond to either (a) tokenized wordpieces in the input or (b) the annotations. Graph edges are used to associate the annotation with the corresponding text spans. Each edge is parametrized with key and value embeddings that are then used in the self-attention formulation; see~\cite{shaw-etal-2019-generating, shaw-etal-2018-self} for details.

\subsubsection*{\bf Copy mechanism:}
The copy mechanism allows the decoder to copy tokens directly from the input (in addition to generating tokens from the vocab). This is particularly useful in our application since we need to copy over the attribute phrase for all intents.

\subsubsection*{\bf Annotation dropout:}
In our model, we copy over text spans and not annotations. This helps the model generalize to unknown attributes which do not come with annotations attached. To further improve generalization, we also randomly drop annotations from the input for a large fraction of examples. This encourages the model to copy over text spans that are not necessarily annotated.

\subsubsection*{\bf Low latency:}
To achieve low latency, we choose a small transformer with 2 encoder layers and a single decoder layer -- we observed only a minor drop in quality when compared with a full transformer. A pictorial representation of our model is presented in Figure~\ref{figure:seq2seq_model}.
\section{Dialog State Tracker (DST)}
\label{section:dst}

\begin{figure}[tb]
    \centering
    \includegraphics[width=\columnwidth]{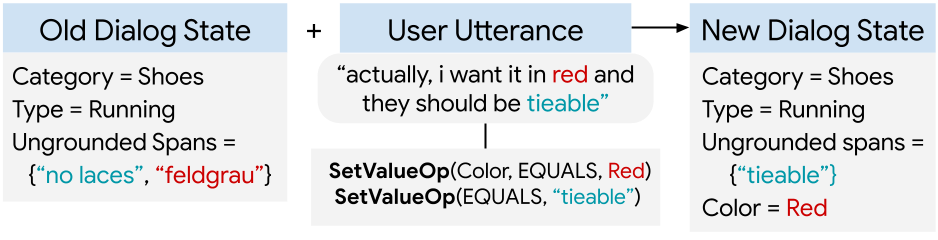}
    \caption{Exemplary dialog state update.}
    \label{figure:dialog_state_update}
\end{figure}

The Dialog State Tracker (DST) manages dialog state. Given an existing dialog state and a set of intents predicted by the CLU for the latest utterance, the DST computes the new dialog state. An example of a DST update using intent operators is shown in Figure~\ref{figure:dialog_state_update}.

\subsection{Intents With Annotated Spans}
\label{subsection:annotated_intents}
If all tags and facets referenced in the intents are recognized by the Shopping schema, then updating the dialog state reduces to a facet (slot) filling problem. For example, if we know from the schema that span \utterance{red} maps to tag \tags{red} in facet \facet{Color}, then the DST can infer the following exemplary rules:
\begin{itemize}
\item \utterance{i only want red} $\rightarrow$ \intentop{SetValueOp}(\facet{Color}, \operator{EQUALS}, \tags{Red}, \intentarg{EXCLUSIVE}) $\Rightarrow$ clear the \facet{Color} facet, then append the tag \tags{Red}
\item \utterance{i also would like to see red ones} $\rightarrow$ \intentop{SetValueOp}(\facet{Color}, \operator{EQUALS}, \tags{Red}, \intentarg{INCLUSIVE}) $\Rightarrow$  append \tags{Red} to the \facet{Color} facet
\item \utterance{just make sure it’s not blue}  $\rightarrow$ \intentop{SetValueOp}(\facet{Color}, \operator{NOT\_EQUALS}, \tags{Blue}) $\Rightarrow$  append \operator{NOT} \tags{Blue} to the \facet{Color} facet
\end{itemize}
As the above examples suggest, the DST update rules vary depending on the intent operator and its arguments ($\langle$facet$\rangle$, $\langle$tag$\rangle$, $\langle$inclusivity$\rangle$, etc.); we discuss intent operators in  Section~\ref{subsection:intent_operators}.

Recall that the CLU outputs a {\em sequence of intents}, the order of which is determined by the user utterance. However, examples below illustrate that this order may be inappropriate for dialog state updates.

\medskip

{\sc Example 1}: \utterance{just make sure they are not blue, but reset other color preferences} may parse into the sequence \{ \intentop{SetValueOp}(\facet{Color}, \operator{NOT\_EQUALS}, \tags{Blue}), \intentop{ClearFacetOp}(\facet{Color}) \} for which the correct ordering is \intentop{ClearFacetOp}(\facet{Color}) followed by \intentop{SetValueOp}(\facet{Color}, \operator{NOT\_EQUALS}, \tags{Blue}).

\smallskip

{\sc Example 2:} \utterance{make sure it's under $\$100$ but reset all other preferences} may parse into the sequence \{ \intentop{SetValueOp}(\facet{Price}, \operator{LESS\_THAN}, \tags{\$100}), \intentop{ClearAllFacetsOp}() \}. The correct ordering is \intentop{ClearAllFacetsOp}() followed by \intentop{SetValueOp}(\facet{Price}, \operator{LESS\_THAN}, \tags{\$100})

\medskip

Examples like those above led us to develop a hand-crafted, domain-independent DST update algorithm in which clear operations are prioritzed; see Figure~\ref{fig:dst_update_algorithm}.

\begin{figure}[h]
    \centering
    \begin{steps}
        \item 
            Apply any facet independent intent operators (\intentop{ClearAllFacetsOp}, \intentop{OrderByOp}).
        \item 
            Group remaining intent operators (\intentop{*ValueOp}, \intentop{*FacetOp}) by facet.
        \item 
            Within each facet grouping, reorder the intent operators such that clear operations appear first.
        \item 
            For each \intentop{SetValueOp}, clear any existing predicates which conflict with the incoming \intentop{SetValueOp} predicate.
        \item 
            Apply remaining intent operators sequentially.
    \end{steps}
    \caption{DST update algorithm.}
    \label{fig:dst_update_algorithm}
\end{figure}

\subsection{Intents With Ungrounded Spans}
\label{subsection:ungrounded_intents}

Ungrounded spans in intents cannot be mapped directly to tags or facets. For an utterance like \utterance{i want shoes that are tieable}, the dialog state update needs an understanding of the relationship between the ungrounded span \utterance{tieable} and various tags and facets already present in dialog state. Note that the existing dialog state itself may contain one or more ungrounded spans as illustrated in Figure~\ref{figure:dialog_state_update}.

To handle ungrounded spans in intents and/or dialog state, we developed a BERT-based deep learned classifier which helps us answer three questions in the context of the product category:
\begin{enumerate}
    \item 
        \texttt{SAME\_TAG(tag1, tag2)} -- do the two tags represent the same concept or attribute? (e.g. \utterance{sleeveless dresses} and \utterance{dresses without sleeves})
    \item 
        \texttt{SAME\_FACET(tag1, tag2)} -- do the two tags belong to the same facet? (e.g \utterance{turquoise shoes} and \utterance{purple shoes})
    \item 
        \texttt{OF\_FACET(tag, facet)} -- does {\texttt tag} belong to {\texttt facet}? (e.g {\texttt tag} = \utterance{turquoise shoes} and {\texttt facet} = \utterance{shoe color})
\end{enumerate}
These three questions are context dependent. For example, an \utterance{ivory dress} and a \utterance{white dress} belong to the same facet (\utterance{color}), whereas an \utterance{ivory figurine} and \utterance{white figurine} do not (\utterance{material} vs \utterance{color}).

For a DST update, we perform pairwise comparisons of every ungrounded span in an intent with every tag, facet and ungrounded span in existing dialog state. These comparisons enable us to apply the DST update algorithm in Figure~\ref{fig:dst_update_algorithm}.

\subsection{Classifier Model}
\label{subsection:classifier_model}
The classifier is trained on examples of the form 
\[ (\texttt{attribute1}, \texttt{attribute2}) \mapsto \texttt{relationships} \]
where {\texttt relationships} is a subset\footnote{Two queries can satisfy multiple labels. For example, any two attributes that satisfy \texttt{SAME\_TAG}, also satisfy \texttt{SAME\_FACET} by definition.} of \{ {\texttt{SAME\_FACET}, \texttt{SAME\_TAG}, \texttt{OF\_FACET}} \}. We train the model on a mix of gold and silver training examples. 

\begin{figure}[tb]
    \centering
    \includegraphics[width=\columnwidth]{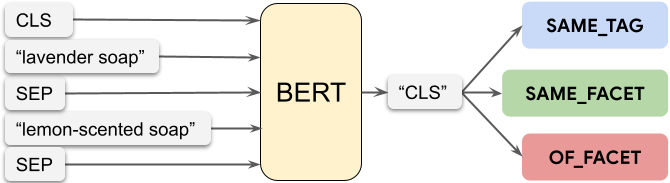}
    \caption{Classifier model multitask architecture.}
    \label{fig:learnt_dst}
\end{figure}

Gold training examples for SAME\_FACET and OF\_FACET are sourced directly from the schema. For example, from the  Shopping schema we know that both \utterance{red} and \utterance{blue} are colors, and therefore we can form three positive examples: (\utterance{red shoes}, \utterance{blue shoes}) $\mapsto$ \{ \texttt{SAME\_FACET} \}, (\utterance{red shoes}, \utterance{shoe color}) $\mapsto$ \{\texttt{OF\_FACET}\}, (\utterance{blue shoes}, \utterance{shoes color}) $\mapsto$ \{ \texttt{OF\_FACET} \}. Negative examples can be formed by sampling tags from different facets in Shopping schema.

We heuristically generate silver examples from multiple sources:
\begin{enumerate}
    \item 
        Category Builder~\cite{mahabal_2018} performs semantic set expansions like \{Ford, Nixon\} $\rightarrow$ \{Obama, Bush, Regan, Trump, \ldots\} and \{Ford, Chevrolet\} $\rightarrow$ \{Jeep, GMC, Tesla, \ldots\}. Starting with seed sets sampled from the Shopping schema, e.g. \{\utterance{red}, \utterance{blue}\}, we construct \texttt{SAME\_FACET} examples using Category Builder; e.g., \{\utterance{red}, \utterance{blue}\} $\mapsto$ \{\utterance{turquoise}, \utterance{feldgrau}, \ldots\}.
    \item 
        Using an internal graph of "Is-A” relationships (e.g. \utterance{red} is a color), constructed using Hearst patterns~\citep{hearst_2012}, we can infer \texttt{SAME\_FACET} relationships by checking whether two tags share several parent node (e.g., \utterance{red} and \utterance{blue} both satisfy "Is-A” color and "Is-A” primary color). We exclude common word parents such as "word” and "thing”.
    \item 
        Internal web synonym services similar to He et al.~\cite{10.1145/2872427.2874816} generate SAME\_TAG examples; e.g., \utterance{big shoes} and \utterance{large shoes}.
\end{enumerate}

Figure \ref{fig:learnt_dst} shows the multitask classifier model. We use a dense projection layer with multiple tasks on top of the BERT output "CLS” token. During training, all tasks are co-trained and all share the same encoder weights, but each task has its own (learned) projection function. For inference, each projection outputs a probability that the input pair of attributes satisfies a particular relationship.

We evaluated our model on crowdworker collected contextual synonyms (\utterance{large shoes} and \utterance{big shoes}) and facet grouping data (\utterance{lavender soap} and \utterance{lemon scented soap}). Model quality was improved by jointly pre-training the encoder on three tasks: (1) masked language modelling (MLM), (2) predicting query parse trees and (3) a general synonym task. The synonym task clearly benefits the very similar SAME\_TAG task. We believe that the MLM and query parse tasks benefit both of the SAME\_FACET and OF\_FACET tasks because both help relate each query token to its root concepts (thereby facilitating broader semantic connections).
\section{Deployment}
\label{section:deployment}

ShopTalk CLU \& DST modules are capable of enabling a conversational shopping experience for any Shopping website or app that builds upon faceted search, for example, using Apache Solr on Lucene~\cite{ApacheLucene,ApacheSolr}. We launched ShopTalk with Google Assistant to leverage pre-existing infrastructure for dialog-based systems. 


In virtual assistants like Google Assistant, Alexa, Siri, Cortana, Dueros (Baidu) and AliGenie, one challenge is multiplexing between multiple backends like Shopping, Maps, Weather and Search for a coherent user experience. Such challenges would not be experienced in the context of a stand-alone Shopping web or app experience. For example, which backend should fulfill \utterance{I want to buy some coffee} -- Maps or Shopping? Another example is this utterance sequence: \utterance{i want to buy shoes} $\rightarrow$ \utterance{size 9} $\rightarrow$ \utterance{waterproof please} $\rightarrow$ \utterance{is it raining?} The last utterance should be tackled by the Weather backend, yet retain Shopping context. For our initial deployment, we relied on heuristics with a plan to build a more robust solution in the future.

For deployment with Google Assistant, we also introduced system prompts such as \utterance{anything else?} and \utterance{got it! What kind of shoe did you have in mind?} to encourage users to utter their preferences. These prompts led to new dialog acts of the forms {\utterance{no}, \utterance{not now}, \utterance{not really}, \utterance{no thank you}, \ldots} or {\utterance{yes}, \utterance{yeah}, \ldots}, which in turn required some straightforward extensions to the CLU \& DST modules to parse such utterances.
\section{Results}
\label{section:results}

{\bf Pre launch metrics:} For end-to-end system evaluation, we asked 500 human raters to interact with a limited deployment of ShopTalk. Raters were instructed to search for products using voice commands, mimicking the behavior of online shoppers. Their first utterance would specify a broad product category, e.g., \utterance{“i want to buy $\langle$\facet{ProductCategory}$\rangle$}. In follow-on utterances, raters were asked to narrow down the set of products using refinement utterances like \utterance{don’t show me pink}. At the end of every turn, the assistant would show an updated list of products. Raters were instructed to continue the conversation until they found a product matching their needs, or if the conversation broke down for any reason. At the end, raters provided a satisfaction rating for the full conversation. Over 70\% raters reported high satisfaction scores.



{\bf Post launch metrics:} After deployment of ShopTalk on Assistant, we observed the following metrics:
\begin{itemize}
\item The number of follow-on Shopping queries increased 9.5x fold. This increase suggests that users actually started refining their queries, engaging in a dialog with the assistant.
\item Average dialog length (number of turns in a conversation) increased by 52\%. This increase suggests the improved capability of parsing (CLU) and dialog state tracking (DST).
\item User engagement with the products that were displayed to them (for example, to inspect detailed product information) in response to their utterances also increased by 52\%. 
\item Very much in line with our pre-launch rater study, over 70\% of the raters using the production system had high user satisfaction score. Even with the increased dialog length, we saw an increase in user satisfaction over the baseline system.
\item Finally, we found that roughly a third of user utterances contain at least one ungrounded span, highlighting how incomplete schemas can be. 
\end{itemize}
\section{Conclusions \& Future Work}
\label{section:conclusions}

Through our deployment of ShopTalk on Google Assistant for conversational shopping, we demonstrated how a real world Faceted Search system can be conversationalized. The underlying schemas for such systems are web-scale (large, complex, incomplete and dynamic), which necessitated clever system design and training data collection techniques. However, further work is needed for a richer experience:

{\bf Support for questions and answers}: Following deployment, we found many users ask questions on the search result page. These questions can be about facets for specific products (\utterance{does it have a usb charging port?} or \utterance{is it compatible with Xbox?}) or about comparing products (\utterance{which phone has the longest battery life?}).

{\bf Better explanation of results}: Further work is needed for (a) annotating every product displayed to the user by justifications, and (b) explaining the set of products as a whole succinctly. For example, users are often surprised if the result set became zero, or remained the same as before; in both cases, an explanation would improve user experience.

{\bf Dialog repair}: (a) Sometimes, the CLU is certain of the intent operator but some argument for that operator is ambiguous. In such cases, instead of the CLU declaring that it failed to parse, dialog repair could be initiated to solicit the ambiguous argument from the user. (b) Consider the utterance sequence \utterance{I want T-shirts} $\rightarrow$ \utterance{something in red} $\rightarrow$ \utterance{show me Nike}. Does the last utterance refer to Nike T-shirts or Nike Shoes? Dialog Repair could help us disambiguate. 

{\bf Out of scope utterances}: Many times, user utterances correspond to functionality that we currently don't support, e.g., comparison of two products. Presently, the CLU and DST fail to parse such utterances. A more graceful handling of such utterances would improve user experience.



{\bf Preference elicitation}: To encourage users to specify their preferences, our system generates a simple 'refinement prompt' like \utterance{anything else?} or \utterance{got it! what kind of $\langle$\productcategory{ProductCategory}$\rangle$ did you have in mind?} This resulted in a wide variety of responses like \utterance{i don't know}, \utterance{stop asking me for specifics}, \utterance{give me a second}, \utterance{i'll just look thank you}, \ldots Further work is needed to respond to such utterances properly and to build richer, more nuanced and context-aware refinement prompts for preference elicitation.

{\bf Guided shopping}: Users often need guidance to explore the product space; they may not know which facets exist, the semantics and trade-offs between different tags in a facet.





\bibliographystyle{ShopTalk-ACM-Reference-Format}
\bibliography{main}


\begin{thebibliography}{38}


\ifx \showCODEN    \undefined \def \showCODEN     #1{\unskip}     \fi
\ifx \showDOI      \undefined \def \showDOI       #1{#1}\fi
\ifx \showISBNx    \undefined \def \showISBNx     #1{\unskip}     \fi
\ifx \showISBNxiii \undefined \def \showISBNxiii  #1{\unskip}     \fi
\ifx \showISSN     \undefined \def \showISSN      #1{\unskip}     \fi
\ifx \showLCCN     \undefined \def \showLCCN      #1{\unskip}     \fi
\ifx \shownote     \undefined \def \shownote      #1{#1}          \fi
\ifx \showarticletitle \undefined \def \showarticletitle #1{#1}   \fi
\ifx \showURL      \undefined \def \showURL       {\relax}        \fi
\providecommand\bibfield[2]{#2}
\providecommand\bibinfo[2]{#2}
\providecommand\natexlab[1]{#1}
\providecommand\showeprint[2][]{arXiv:#2}

\bibitem[\protect\citeauthoryear{{Apache Software Foundation}}{{Apache Software
  Foundation}}{[n.d.]a}]%
        {ApacheLucene}
\bibfield{author}{\bibinfo{person}{{Apache Software Foundation}}.}
  \bibinfo{year}{[n.d.]}\natexlab{a}.
\newblock \bibinfo{title}{{Apache Lucene}}.
\newblock \bibinfo{howpublished}{\url{https://lucene.apache.org/}}.
\newblock
\newblock
\shownote{[Online; accessed 3-February-2021].}


\bibitem[\protect\citeauthoryear{{Apache Software Foundation}}{{Apache Software
  Foundation}}{[n.d.]b}]%
        {ApacheSolr}
\bibfield{author}{\bibinfo{person}{{Apache Software Foundation}}.}
  \bibinfo{year}{[n.d.]}\natexlab{b}.
\newblock \bibinfo{title}{{Apache Solr}}.
\newblock \bibinfo{howpublished}{\url{https://lucene.apache.org/solr/}}.
\newblock
\newblock
\shownote{[Online; accessed 3-February-2021].}


\bibitem[\protect\citeauthoryear{Budzianowski, Wen, Tseng, Casanueva, Ultes,
  Ramadan, and Ga{\v{s}}i{\'c}}{Budzianowski et~al\mbox{.}}{2018}]%
        {Budzianowski2018}
\bibfield{author}{\bibinfo{person}{Pawe{\l} Budzianowski},
  \bibinfo{person}{Tsung-Hsien Wen}, {et~al\mbox{.}}}
  \bibinfo{year}{2018}\natexlab{}.
\newblock \showarticletitle{{M}ulti{WOZ} - A Large-Scale Multi-Domain
  {W}izard-of-{O}z Dataset for Task-Oriented Dialogue Modelling}. In
  \bibinfo{booktitle}{\emph{EMNLP 2018}}.
\newblock


\bibitem[\protect\citeauthoryear{Chao and Lane}{Chao and Lane}{2019}]%
        {Chao2019}
\bibfield{author}{\bibinfo{person}{Guan-Lin Chao} {and} \bibinfo{person}{Ian
  Lane}.} \bibinfo{year}{2019}\natexlab{}.
\newblock \showarticletitle{{BERT-DST}: Scalable End-to-End Dialogue State
  Tracking with Bidirectional Encoder Representations from Transformer}. In
  \bibinfo{booktitle}{\emph{INTERSPEECH}}.
\newblock


\bibitem[\protect\citeauthoryear{Devlin, Chang, Lee, and Toutanova}{Devlin
  et~al\mbox{.}}{2019}]%
        {devlin-etal-2019-bert}
\bibfield{author}{\bibinfo{person}{Jacob Devlin}, \bibinfo{person}{Ming-Wei
  Chang}, {et~al\mbox{.}}} \bibinfo{year}{2019}\natexlab{}.
\newblock \showarticletitle{{BERT}: Pre-training of Deep Bidirectional
  Transformers for Language Understanding}. In \bibinfo{booktitle}{\emph{NAACL
  2019}}.
\newblock


\bibitem[\protect\citeauthoryear{Dong and Lapata}{Dong and Lapata}{2016}]%
        {dong-lapata-2016-language}
\bibfield{author}{\bibinfo{person}{Li Dong} {and} \bibinfo{person}{Mirella
  Lapata}.} \bibinfo{year}{2016}\natexlab{}.
\newblock \showarticletitle{Language to Logical Form with Neural Attention}. In
  \bibinfo{booktitle}{\emph{ACL 2016}}.
\newblock


\bibitem[\protect\citeauthoryear{Dong and Lapata}{Dong and Lapata}{2018}]%
        {dong-lapata-2018-coarse}
\bibfield{author}{\bibinfo{person}{Li Dong} {and} \bibinfo{person}{Mirella
  Lapata}.} \bibinfo{year}{2018}\natexlab{}.
\newblock \showarticletitle{Coarse-to-Fine Decoding for Neural Semantic
  Parsing}. In \bibinfo{booktitle}{\emph{ACL 2018}}.
\newblock


\bibitem[\protect\citeauthoryear{El~Asri, Schulz, Sharma, Zumer, Harris, Fine,
  Mehrotra, and Suleman}{El~Asri et~al\mbox{.}}{2017}]%
        {Asri2017}
\bibfield{author}{\bibinfo{person}{Layla El~Asri}, \bibinfo{person}{Hannes
  Schulz}, {et~al\mbox{.}}} \bibinfo{year}{2017}\natexlab{}.
\newblock \showarticletitle{{F}rames: a corpus for adding memory to
  goal-oriented dialogue systems}. In \bibinfo{booktitle}{\emph{SIGDIAL 2017}}.
\newblock


\bibitem[\protect\citeauthoryear{Gao, Sethi, Agarwal, Chung, and
  Hakkani-Tur}{Gao et~al\mbox{.}}{2019}]%
        {Gao2019}
\bibfield{author}{\bibinfo{person}{Shuyang Gao}, \bibinfo{person}{Abhishek
  Sethi}, {et~al\mbox{.}}} \bibinfo{year}{2019}\natexlab{}.
\newblock \showarticletitle{Dialog State Tracking: A Neural Reading
  Comprehension Approach}. In \bibinfo{booktitle}{\emph{SIGDIAL 2019}}.
\newblock


\bibitem[\protect\citeauthoryear{Goel, Paul, and Hakkani-Tür}{Goel
  et~al\mbox{.}}{2019}]%
        {Goel2019}
\bibfield{author}{\bibinfo{person}{Rahul Goel}, \bibinfo{person}{Shachi Paul},
  {et~al\mbox{.}}} \bibinfo{year}{2019}\natexlab{}.
\newblock \showarticletitle{{HyST: A Hybrid Approach for Flexible and Accurate
  Dialogue State Tracking}}. In \bibinfo{booktitle}{\emph{Proc. Interspeech
  2019}}.
\newblock


\bibitem[\protect\citeauthoryear{Gong, Luo, Zhu, Ou, Li, Zhu, Zhu, Duan, and
  Chen}{Gong et~al\mbox{.}}{2019}]%
        {DBLP:conf/aaai/GongLZOLZZDC19}
\bibfield{author}{\bibinfo{person}{Yu Gong}, \bibinfo{person}{Xusheng Luo},
  {et~al\mbox{.}}} \bibinfo{year}{2019}\natexlab{}.
\newblock \showarticletitle{Deep Cascade Multi-Task Learning for Slot Filling
  in Online Shopping Assistant}. In \bibinfo{booktitle}{\emph{AAAI 2019}}.
  \bibinfo{pages}{6465--6472}.
\newblock


\bibitem[\protect\citeauthoryear{{Google Inc}}{{Google Inc}}{2021}]%
        {GoogleProductTaxonomy}
\bibfield{author}{\bibinfo{person}{{Google Inc}}.}
  \bibinfo{year}{2021}\natexlab{}.
\newblock \bibinfo{title}{{Google Product Taxonomy}}.
\newblock
  \bibinfo{howpublished}{\url{https://www.google.com/basepages/producttype/taxonomy.en-US.txt}}.
\newblock
\newblock
\shownote{[Online; accessed 3-February-2021].}


\bibitem[\protect\citeauthoryear{He, Chakrabarti, Cheng, and Tylenda}{He
  et~al\mbox{.}}{2016}]%
        {10.1145/2872427.2874816}
\bibfield{author}{\bibinfo{person}{Yeye He}, \bibinfo{person}{Kaushik
  Chakrabarti}, {et~al\mbox{.}}} \bibinfo{year}{2016}\natexlab{}.
\newblock \showarticletitle{Automatic Discovery of Attribute Synonyms Using
  Query Logs and Table Corpora} \emph{(\bibinfo{series}{WWW '16})}.
\newblock


\bibitem[\protect\citeauthoryear{Hearst}{Hearst}{1992}]%
        {hearst_2012}
\bibfield{author}{\bibinfo{person}{Marti~A. Hearst}.}
  \bibinfo{year}{1992}\natexlab{}.
\newblock \showarticletitle{Automatic Acquisition of Hyponyms from Large Text
  Corpora}. In \bibinfo{booktitle}{\emph{{COLING} 1992 Volume 2: The 15th
  {I}nternational {C}onference on {C}omputational {L}inguistics}}.
\newblock


\bibitem[\protect\citeauthoryear{Henderson, Thomson, and Young}{Henderson
  et~al\mbox{.}}{2014}]%
        {Henderson2014}
\bibfield{author}{\bibinfo{person}{Matthew Henderson}, \bibinfo{person}{Blaise
  Thomson}, {et~al\mbox{.}}} \bibinfo{year}{2014}\natexlab{}.
\newblock \showarticletitle{Word-Based Dialog State Tracking with Recurrent
  Neural Networks}. In \bibinfo{booktitle}{\emph{SIGDIAL 2014}}.
\newblock


\bibitem[\protect\citeauthoryear{Iyer, Konstas, Cheung, Krishnamurthy, and
  Zettlemoyer}{Iyer et~al\mbox{.}}{2017}]%
        {iyer-etal-2017-learning}
\bibfield{author}{\bibinfo{person}{Srinivasan Iyer}, \bibinfo{person}{Ioannis
  Konstas}, {et~al\mbox{.}}} \bibinfo{year}{2017}\natexlab{}.
\newblock \showarticletitle{Learning a Neural Semantic Parser from User
  Feedback}. In \bibinfo{booktitle}{\emph{ACL 2017}}.
\newblock


\bibitem[\protect\citeauthoryear{Jia and Liang}{Jia and Liang}{2016}]%
        {jia-liang-2016-data}
\bibfield{author}{\bibinfo{person}{Robin Jia} {and} \bibinfo{person}{Percy
  Liang}.} \bibinfo{year}{2016}\natexlab{}.
\newblock \showarticletitle{Data Recombination for Neural Semantic Parsing}. In
  \bibinfo{booktitle}{\emph{ACL 2016}}.
\newblock


\bibitem[\protect\citeauthoryear{Lee and Eskenazi}{Lee and Eskenazi}{2013}]%
        {Lee2013}
\bibfield{author}{\bibinfo{person}{Sungjin Lee} {and} \bibinfo{person}{Maxine
  Eskenazi}.} \bibinfo{year}{2013}\natexlab{}.
\newblock \showarticletitle{Recipe For Building Robust Spoken Dialog State
  Trackers: Dialog State Tracking Challenge System Description}. In
  \bibinfo{booktitle}{\emph{SIGDIAL 2013}}.
\newblock


\bibitem[\protect\citeauthoryear{Lin, Bogin, Neumann, Berant, and Gardner}{Lin
  et~al\mbox{.}}{2019}]%
        {lin2019grammarbased}
\bibfield{author}{\bibinfo{person}{Kevin Lin}, \bibinfo{person}{Ben Bogin},
  {et~al\mbox{.}}} \bibinfo{year}{2019}\natexlab{}.
\newblock \bibinfo{title}{Grammar-based Neural Text-to-SQL Generation}.
\newblock
\newblock
\showeprint[arxiv]{1905.13326}~[cs.CL]


\bibitem[\protect\citeauthoryear{Mahabal, Roth, and Mittal}{Mahabal
  et~al\mbox{.}}{2018}]%
        {mahabal_2018}
\bibfield{author}{\bibinfo{person}{Abhijit Mahabal}, \bibinfo{person}{Dan
  Roth}, {et~al\mbox{.}}} \bibinfo{year}{2018}\natexlab{}.
\newblock \showarticletitle{Robust Handling of Polysemy via Sparse
  Representations}. In \bibinfo{booktitle}{\emph{Proceedings of the Seventh
  Joint Conference on Lexical and Computational Semantics}}.
\newblock


\bibitem[\protect\citeauthoryear{Mrk{\v{s}}i{\'c}, {\'O}~S{\'e}aghdha, Wen,
  Thomson, and Young}{Mrk{\v{s}}i{\'c} et~al\mbox{.}}{2017}]%
        {Mrksic2016}
\bibfield{author}{\bibinfo{person}{Nikola Mrk{\v{s}}i{\'c}},
  \bibinfo{person}{Diarmuid {\'O}~S{\'e}aghdha}, {et~al\mbox{.}}}
  \bibinfo{year}{2017}\natexlab{}.
\newblock \showarticletitle{Neural Belief Tracker: Data-Driven Dialogue State
  Tracking}. In \bibinfo{booktitle}{\emph{ACL 2017}}.
\newblock


\bibitem[\protect\citeauthoryear{Perez and Liu}{Perez and Liu}{2017}]%
        {Perez2016}
\bibfield{author}{\bibinfo{person}{Julien Perez} {and} \bibinfo{person}{Fei
  Liu}.} \bibinfo{year}{2017}\natexlab{}.
\newblock \showarticletitle{Dialog state tracking, a machine reading approach
  using Memory Network}. In \bibinfo{booktitle}{\emph{Proceedings of the 15th
  Conference of the {E}uropean Chapter of the Association for Computational
  Linguistics: Volume 1, Long Papers}}.
\newblock


\bibitem[\protect\citeauthoryear{Ramadan, Budzianowski, and
  Ga{\v{s}}i{\'c}}{Ramadan et~al\mbox{.}}{2018}]%
        {Ramadan2018}
\bibfield{author}{\bibinfo{person}{Osman Ramadan}, \bibinfo{person}{Pawe{\l}
  Budzianowski}, {et~al\mbox{.}}} \bibinfo{year}{2018}\natexlab{}.
\newblock \showarticletitle{Large-Scale Multi-Domain Belief Tracking with
  Knowledge Sharing}. In \bibinfo{booktitle}{\emph{ACL 2018}}.
\newblock


\bibitem[\protect\citeauthoryear{Rastogi, Zang, Sunkara, Gupta, and
  Khaitan}{Rastogi et~al\mbox{.}}{2020}]%
        {Rastogi2019}
\bibfield{author}{\bibinfo{person}{Abhinav Rastogi}, \bibinfo{person}{Xiaoxue
  Zang}, {et~al\mbox{.}}} \bibinfo{year}{2020}\natexlab{}.
\newblock \showarticletitle{Towards Scalable Multi-Domain Conversational
  Agents: The Schema-Guided Dialogue Dataset}.
\newblock \bibinfo{journal}{\emph{AAAI 2020}} (\bibinfo{year}{2020}).
\newblock


\bibitem[\protect\citeauthoryear{Ren, Xie, Chen, and Yu}{Ren
  et~al\mbox{.}}{2018}]%
        {Ren2018}
\bibfield{author}{\bibinfo{person}{Liliang Ren}, \bibinfo{person}{Kaige Xie},
  {et~al\mbox{.}}} \bibinfo{year}{2018}\natexlab{}.
\newblock \showarticletitle{Towards Universal Dialogue State Tracking}. In
  \bibinfo{booktitle}{\emph{EMNLP 2018}}.
\newblock


\bibitem[\protect\citeauthoryear{Shaw, Massey, Chen, Piccinno, and Altun}{Shaw
  et~al\mbox{.}}{2019}]%
        {shaw-etal-2019-generating}
\bibfield{author}{\bibinfo{person}{Peter Shaw}, \bibinfo{person}{Philip
  Massey}, {et~al\mbox{.}}} \bibinfo{year}{2019}\natexlab{}.
\newblock \showarticletitle{Generating Logical Forms from Graph Representations
  of Text and Entities}. In \bibinfo{booktitle}{\emph{ACL 2019}}.
\newblock


\bibitem[\protect\citeauthoryear{Shaw, Uszkoreit, and Vaswani}{Shaw
  et~al\mbox{.}}{2018}]%
        {shaw-etal-2018-self}
\bibfield{author}{\bibinfo{person}{Peter Shaw}, \bibinfo{person}{Jakob
  Uszkoreit}, {et~al\mbox{.}}} \bibinfo{year}{2018}\natexlab{}.
\newblock \showarticletitle{Self-Attention with Relative Position
  Representations}. In \bibinfo{booktitle}{\emph{NAACL 2018}}.
\newblock


\bibitem[\protect\citeauthoryear{Thomson and Young}{Thomson and Young}{2010}]%
        {Thomson2010}
\bibfield{author}{\bibinfo{person}{Blaise Thomson} {and} \bibinfo{person}{Steve
  Young}.} \bibinfo{year}{2010}\natexlab{}.
\newblock \showarticletitle{{Bayesian update of dialogue state: A POMDP
  framework for spoken dialogue systems}}.
\newblock \bibinfo{journal}{\emph{{Computer Speech and Language}}}
  \bibinfo{volume}{24}, \bibinfo{number}{4} (\bibinfo{date}{March}
  \bibinfo{year}{2010}).
\newblock


\bibitem[\protect\citeauthoryear{Vaswani, Shazeer, Parmar, Uszkoreit, Jones,
  Gomez, Kaiser, and Polosukhin}{Vaswani et~al\mbox{.}}{2017}]%
        {NIPS2017_3f5ee243}
\bibfield{author}{\bibinfo{person}{Ashish Vaswani}, \bibinfo{person}{Noam
  Shazeer}, {et~al\mbox{.}}} \bibinfo{year}{2017}\natexlab{}.
\newblock \showarticletitle{Attention is All you Need}. In
  \bibinfo{booktitle}{\emph{Advances in Neural Information Processing
  Systems}}, Vol.~\bibinfo{volume}{30}.
\newblock


\bibitem[\protect\citeauthoryear{Wang and Lemon}{Wang and Lemon}{2013}]%
        {Wang2013}
\bibfield{author}{\bibinfo{person}{Zhuoran Wang} {and} \bibinfo{person}{Oliver
  Lemon}.} \bibinfo{year}{2013}\natexlab{}.
\newblock \showarticletitle{A Simple and Generic Belief Tracking Mechanism for
  the Dialog State Tracking Challenge: On the believability of observed
  information}. In \bibinfo{booktitle}{\emph{SIGDIAL 2013}}.
\newblock


\bibitem[\protect\citeauthoryear{Weld, Huang, Long, Poon, and Han}{Weld
  et~al\mbox{.}}{2021}]%
        {weld2021survey}
\bibfield{author}{\bibinfo{person}{H. Weld}, \bibinfo{person}{X. Huang},
  {et~al\mbox{.}}} \bibinfo{year}{2021}\natexlab{}.
\newblock \bibinfo{title}{A survey of joint intent detection and slot-filling
  models in natural language understanding}.
\newblock
\newblock
\showeprint[arxiv]{2101.08091}~[cs.CL]


\bibitem[\protect\citeauthoryear{Wen, Vandyke, Mrk{\v{s}}i{\'c},
  Ga{\v{s}}i{\'c}, Rojas-Barahona, Su, Ultes, and Young}{Wen
  et~al\mbox{.}}{2017}]%
        {Wen2016}
\bibfield{author}{\bibinfo{person}{Tsung-Hsien Wen}, \bibinfo{person}{David
  Vandyke}, {et~al\mbox{.}}} \bibinfo{year}{2017}\natexlab{}.
\newblock \showarticletitle{A Network-based End-to-End Trainable Task-oriented
  Dialogue System}. In \bibinfo{booktitle}{\emph{Proceedings of the 15th
  Conference of the {E}uropean Chapter of the Association for Computational
  Linguistics}}.
\newblock


\bibitem[\protect\citeauthoryear{Wu, Madotto, Hosseini-Asl, Xiong, Socher, and
  Fung}{Wu et~al\mbox{.}}{2019}]%
        {Wu2019}
\bibfield{author}{\bibinfo{person}{Chien-Sheng Wu}, \bibinfo{person}{Andrea
  Madotto}, {et~al\mbox{.}}} \bibinfo{year}{2019}\natexlab{}.
\newblock \showarticletitle{Transferable Multi-Domain State Generator for
  Task-Oriented Dialogue Systems}. In \bibinfo{booktitle}{\emph{ACL 2019}}.
\newblock


\bibitem[\protect\citeauthoryear{Xu, Wang, Mao, Jiang, and Lan}{Xu
  et~al\mbox{.}}{2019}]%
        {xu-etal-2019-scaling}
\bibfield{author}{\bibinfo{person}{Huimin Xu}, \bibinfo{person}{Wenting Wang},
  {et~al\mbox{.}}} \bibinfo{year}{2019}\natexlab{}.
\newblock \showarticletitle{Scaling up Open Tagging from Tens to Thousands:
  Comprehension Empowered Attribute Value Extraction from Product Title}. In
  \bibinfo{booktitle}{\emph{ACL 2019}}.
\newblock


\bibitem[\protect\citeauthoryear{Xu and Hu}{Xu and Hu}{2018}]%
        {Xu2018}
\bibfield{author}{\bibinfo{person}{Puyang Xu} {and} \bibinfo{person}{Qi Hu}.}
  \bibinfo{year}{2018}\natexlab{}.
\newblock \showarticletitle{An End-to-end Approach for Handling Unknown Slot
  Values in Dialogue State Tracking}. In \bibinfo{booktitle}{\emph{ACL 2018}}.
\newblock


\bibitem[\protect\citeauthoryear{Yan, Duan, Chen, Zhou, Zhou, and Li}{Yan
  et~al\mbox{.}}{2017}]%
        {yan2017}
\bibfield{author}{\bibinfo{person}{Zhao Yan}, \bibinfo{person}{Nan Duan},
  {et~al\mbox{.}}} \bibinfo{year}{2017}\natexlab{}.
\newblock \showarticletitle{Building Task-Oriented Dialogue Systems for Online
  Shopping}. In \bibinfo{booktitle}{\emph{AAAI'17}}.
  \bibinfo{pages}{4618–4625}.
\newblock


\bibitem[\protect\citeauthoryear{Yang, Gong, and Chen}{Yang
  et~al\mbox{.}}{2018}]%
        {Yang2018}
\bibfield{author}{\bibinfo{person}{Yunlun Yang}, \bibinfo{person}{Yu Gong},
  {et~al\mbox{.}}} \bibinfo{year}{2018}\natexlab{}.
\newblock \bibinfo{title}{Query Tracking for E-commerce Conversational Search:
  A Machine Comprehension Perspective}.
\newblock
\newblock
\showeprint[arxiv]{1810.03274}~[cs.CL]


\bibitem[\protect\citeauthoryear{Zhang, Yu, Er, Shim, Xue, Lin, Shi, Xiong,
  Socher, and Radev}{Zhang et~al\mbox{.}}{2019}]%
        {zhang-etal-2019-editing}
\bibfield{author}{\bibinfo{person}{Rui Zhang}, \bibinfo{person}{Tao Yu},
  {et~al\mbox{.}}} \bibinfo{year}{2019}\natexlab{}.
\newblock \showarticletitle{Editing-Based {SQL} Query Generation for
  Cross-Domain Context-Dependent Questions}. In
  \bibinfo{booktitle}{\emph{EMNLP-IJCNLP 2019}}.
\newblock


\end{thebibliography}


\end{document}